\documentclass{article}

\usepackage{arxiv}

\usepackage[utf8]{inputenc}
\usepackage[T1]{fontenc}
\usepackage{hyperref}
\usepackage{url}
\usepackage{booktabs}
\usepackage{wrapfig}
\usepackage{caption}
\usepackage{subcaption}
\usepackage[export]{adjustbox}
\usepackage{graphicx}
\usepackage{multirow}
\usepackage{amsmath}
\usepackage{amssymb}
\usepackage{xcolor}
\usepackage{tikz}
\usepackage{siunitx}
\usepackage{comment}
\usepackage{array}
\usepackage{pifont}
\usepackage{microtype}

\newcommand{\net}{TriDerm}

\newcommand{\secref}[1]{Sec.~\ref{#1}}
\renewcommand{\eqref}[1]{Eq.~(\ref{#1})}
\newcommand{\figref}[1]{Fig.~\ref{#1}}
\newcommand{\tabref}[1]{Tab.~\ref{#1}}

\renewcommand{\headeright}{}
\renewcommand{\undertitle}{}
\renewcommand{\shorttitle}{Assessing Multimodal Chronic Wound Embeddings with Expert Triplet Agreement}

\title{Assessing Multimodal Chronic Wound Embeddings with Expert Triplet Agreement}
\author{
\parbox{0.97\textwidth}{\centering
Fabian Kabus$^{1}$\thanks{These authors contributed equally.}, Julia Hindel$^{2}$\footnotemark[1], Jelena Bratuli\'c$^{2}$, Meropi Karakioulaki$^{3}$,\\
Ayush Gupta$^{2}$, Cristina Has$^{3}$, Thomas Brox$^{2}$, Abhinav Valada$^{2}$, Harald Binder$^{1}$\\[0.55em]
{\mdseries
$^{1}$Institute of Medical Biometry and Statistics (IMBI), Medical Faculty and Medical Center, University of Freiburg\\
$^{2}$Department of Computer Science, Faculty of Engineering, University of Freiburg\\
$^{3}$Department of Dermatology, Medical Faculty and Medical Center, University of Freiburg\\[0.75em]
\texttt{\{fabian.kabus, harald.binder\}@uniklinik-freiburg.de}\\
\texttt{\{hindel, bratulic, guptay, brox, valada\}@cs.uni-freiburg.de}\\
\texttt{\{meropi.karakioulaki, cristina.has\}@uniklinik-freiburg.de}
}
}
}

\begin{document}
\maketitle

\begin{abstract}







Recessive dystrophic epidermolysis bullosa (RDEB) is a rare genetic skin disorder for which clinicians greatly benefit from finding similar cases using images and clinical text. However, off‑the‑shelf foundation models do not reliably capture clinically meaningful features for this heterogeneous, long‑tail disease, and structured measurement of agreement with experts is challenging. To address these gaps, we propose evaluating embedding spaces with expert ordinal comparisons (triplet judgments), which are fast to collect and encode implicit clinical similarity knowledge. We further introduce \net{}, a multimodal framework that learns interpretable wound representations from small cohorts by integrating wound imagery, boundary masks, and expert reports. On the vision side, \net{} adapts visual foundation models to RDEB using wound‑level attention pooling and non‑contrastive representation learning. For text, we prompt large language models with comparison queries and recover medically meaningful representations via soft ordinal embeddings (SOE). We show that visual and textual modalities capture complementary aspects of wound phenotype, and that fusing both modalities yields 73.5\% agreement with experts, outperforming the best off‑the‑shelf single‑modality foundation model by over 5.6 percentage points. We make the expert annotation tool, model code and representative dataset samples publicly available.

\end{abstract}


\section{Introduction}

Large-scale dermatology datasets and foundation models have driven substantial progress in skin disease classification and retrieval~\cite{yan2025derm1m,yan2026visionlanguage}. However, prior work targets broad taxonomic hierarchies, leaving rare diseases in the long tail of pretraining distributions. Furthermore, current approaches do not capture fine-grained phenotypic structure within specific rare diseases. This is a critical gap, as identifying similar cases and their treatment outcomes is particularly relevant for clinical decision-making in rare diseases.

Recessive dystrophic epidermolysis bullosa (RDEB) exemplifies this challenge. This rare, chronic blistering disease affects approximately 3.5--20.4 per million individuals~\cite{bardhan2020epidermolysis}, causing profound skin fragility, chronic non-healing wounds, recurrent infections, and elevated squamous cell carcinoma risk. The typical assessment of patients comprises wound images as well as text descriptions. Yet the substantial heterogeneity in wound appearance and severity remains poorly understood, limiting both clinical decisions and data-driven research.

This challenge motivates embedding-based retrieval, in which wound similarity can be measured directly within a learned representation space. However, validating these similarities requires a structured, expert‑anchored assessment to ensure clinical fidelity. Consequently, we introduce triplet judgments of a dermatology expert as a reference for assessing embeddings of image and text data. Given two reference wound images, the medical expert indicates for the third image to which reference image it is most similar in appearance. Unlike categorical labels, triplets encode ordinal similarity structure, representing graded and overlapping phenotypic relationships that resist discrete categorization.


To improve embeddings, we propose \net{}, which adapts vision foundation models via an attention‑pooling module trained with a self‑supervised objective on annotated wound regions. For text processing, we propose a synthetic-expert protocol that queries LLMs with wound descriptions and extracts triplet judgments, mirroring the structure of the expert judgment, to adapt LLMs in this small data regime. Since vision and text encoders may capture complementary aspects of wound phenotypes, we also evaluate multimodal fusion strategies that combine their representations.

Our contributions are as follows:
\begin{enumerate}
    \item We introduce a triplet paradigm to evaluate medical reasoning of visual, textual, and multimodal representations in a rare-disease setting.
     \item Our method \net{}~introduces wound-level attention pooling with non-contrastive learning to adapt visual foundation models to RDEB-specific characteristics.
     \item We propose triplet questions to discover clinically meaningful embeddings via soft ordinal embedding (SOE) from only 53 unlabeled text descriptions of RDEB wounds.
    \item We publicly release the expert annotation tool, model code and representative dataset samples upon acceptance.
\end{enumerate}



\section{Method}\label{sec:method}

\subsection{Dataset and Expert Annotations}
We collected 53~wound camera images from 21~RDEB patients at a University Medical Center unit specialized on RDEB. A dermatologist manually delineated separate wounds in each image and provided free-text descriptions characterizing each wound's clinical appearance.
Each RDEB wound was annotated separately, yielding 120 labeled wounds, with images containing between 1--16 wounds across varying sizes and anatomical sites. The images were collected during routine clinical care by treating clinicians using standard camera equipment. As is typical in real-world settings, no uniform photography protocol (e.g., fixed viewpoint, distance, lighting, or positioning) was mandated, resulting in natural variation in image capture conditions. 
The text descriptions capture features routinely assessed in wound care, such as granulation tissue level, fibrin coverage, inflammation, depth, erythema, border morphology (hyperkeratotic, fibrotic, scarred), and surface characteristics such as hemorrhagic crusts.
In addition, a dermatologist performed triplet judgments: given two reference wound images, the dermatologist indicated which of the reference images a third image is more similar to. This was performed for 513 randomly selected triplets.

\subsection{Embedding Methods}
\begin{figure}[t]
\centering
\includegraphics[width=0.9\textwidth]{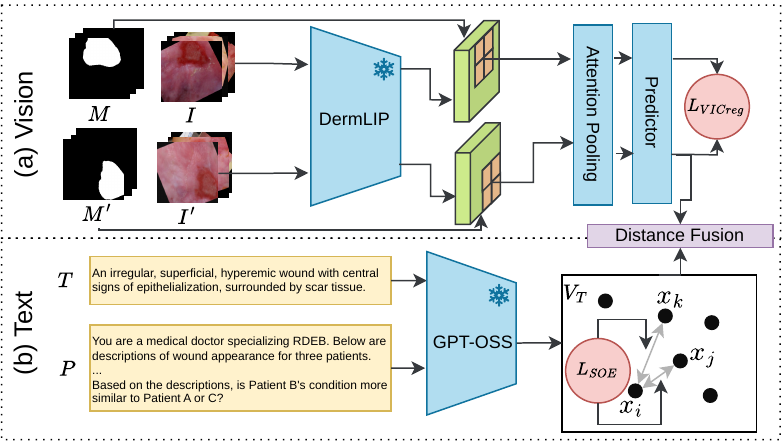}
\caption{\net~leverages triplet questions to obtain embeddings via SOE that mirror medical expert similarity structure. To improve visual wound representations, our network combines non-contrastive representation learning with attention pooling. Uncertainty-aware late fusion is performed on distance matrices.}
\label{fig:model}
\end{figure}

\paragraph{Visual Representation:}
Skin foundation models, DermLIP~\cite{dermlip2024} and DermFM-Zero~\cite{yan2026visionlanguage}, demonstrate strong capabilities 
in wound semantics, yet often lack precise differentiation of wound severity and fine-grained degrees of tissue damage. Consequently, we propose a wound-wise representation learning framework as shown in \figref{fig:model}a. Given a camera image of a wound, we first produce two independent augmented views ($I$, $I'$). Both views are passed through a frozen DermLIP-ViT-B~\cite{dermlip2024} backbone, yielding feature maps of shape $\mathbb{R}^{B \times C \times H \times W}$. Wound-specific 
feature vectors are then extracted by sampling all spatial locations within the wound region, resulting in a feature tensor of shape $\mathbb{R}^{B \times N \times C}$, where $N$ denotes the number of sampled wound tokens. To aggregate these features, we apply attention pooling per wound, implemented as a two-layer MLP with interleaved $\tanh$ activations, producing an attention mask of shape $\mathbb{R}^{B \times N}$. The attended features are subsequently encoded by a lightweight predictor head comprising a single linear layer followed by Layer Normalization to produce the final wound-wise representations $F_V$. We train the attention pooling and predictor head using the VICReg loss~\cite{bardes2021vicreg}, a non-contrastive self-supervised objective that operates on paired augmented features $(F_V, F_V')$. Unlike contrastive methods, VICReg does not require negative samples, making it particularly well-suited for small datasets. The loss is defined as

\begin{equation}
    \mathcal{L}_{\text{VICReg}} = \lambda \cdot s(F_V, F_V') + \mu \cdot v(F_V, F_V') + \nu \cdot c(F_V, F_V'),
\end{equation}

\noindent where $s$ is an invariance term that enforces consistency between the two features by minimizing their mean squared error; $v$ is a variance term that prevents representational collapse by enforcing a minimum standard deviation across the batch; and $c$ is a covariance 
term that decorrelates the embedding dimensions to reduce information redundancy. We ablate the effect of alternative representation learning objectives in \secref{sec:ab}. For evaluation, we obtain a single feature embedding per image by computing the mean across all wound-wise representations, which we find sufficient in practice.

\paragraph{Language Representation:}
Each wound image is accompanied by expert-written descriptions ($T$) that characterize the clinical appearance, including granulation tissue, fibrin coverage, inflammation, depth, and border morphology. These free-text descriptions serve as input for text-based similarity, as shown in \figref{fig:model}b.
Rather than directly probing LLM activations---which may not capture clinically meaningful similarity---we treat the LLM as a synthetic expert. Given three wound descriptions (Cases i, j, k), we prompt the model with system prompt $P$: ``Is wound i more similar to wound j or to wound k?'' $P$ establishes an RDEB specialist persona to ground the comparison in disease-specific reasoning. We sample triplets uniformly from all possible combinations and collect the LLM's binary responses.
These synthetic triplet judgments define ordinal constraints on wound similarity. We recover a continuous embedding by fitting Soft Ordinal Embedding (SOE)~\cite{terada2014local} to the LLM-derived triplets. SOE minimizes a hinge loss that enforces $d(F_T^{(i)}, F_T^{(j)}) < d(F_T^{(i)}, F_T^{(k)})$ when the anchor $i$ is judged closer to $j$ than to $k$:
\begin{equation}
    \mathcal{L}_{\text{SOE}} = \sum_{(i,j,k)} \max\bigl(0,\; \delta + \|F_T^{(i)} - F_T^{(j)}\| - \|F_T^{(i)} - F_T^{(k)}\|\bigr),
\end{equation}
where $\delta$ is a margin hyperparameter. The resulting embedding $F_T$ captures the LLM's expressed medical reasoning rather than its internal activation geometry.

\paragraph{Distance Fusion:}

Vision and text embeddings may capture complementary aspects of wounds. We evaluate two fusion strategies that operate on pairwise distance matrices without requiring additional training. 
Both approaches rely on min-max normalized distance matrices: $\tilde{D}_V$ and $\tilde{D}_T$ denotes the normalized vision and text distance matrices, respectively.

\textbf{Uncertainty-aware fusion:} Modality reliability varies per sample. We weight each modality by its confidence, measured as the variance $\sigma^2_{m,i}$ of distances from sample $i$ to all other samples, where high variance signals clear separation and low variance (equidistance) indicates uncertainty:
\begin{equation}
    w_i = \frac{\alpha \cdot \sigma^2_{V,i}}{\alpha \cdot \sigma^2_{V,i} + (1-\alpha) \cdot \sigma^2_{T,i}},
\end{equation}
where $\alpha=0.7$ to account for the vision model's empirically stronger performance. The fused distance estimates per-sample are defined as:
\begin{equation}
    D_{\text{fused}}(i,j) = \bar{w}_{ij} \cdot \tilde{D}_V(i,j) + (1 - \bar{w}_{ij}) \cdot \tilde{D}_T(i,j), \quad \bar{w}_{ij} = \tfrac{1}{2}(w_i + w_j).
\end{equation}

\textbf{Similarity-based fusion:} We convert distances to similarities $S_m = 1 - \tilde{D}_m$ and compute $S_{\text{fused}} = S_V \odot S_T$, yielding $D_{\text{fused}} = 1 - S_{\text{fused}}$. Since vision and text rely on different features, their errors tend to be weakly correlated. Consequently, agreement filters spurious matches that arise from a single modality.

\section{Experiments}

\subsection{Evaluation Metrics}
We evaluate embedding quality via triplet agreements from a dermatologist. Following \secref{sec:method}, we define the triplet margin $\Delta_t = d(i,k) - d(i,j)$, where a triplet is considered correct when $\Delta_t > 0$ for expert-annotated triplets. To account for unequal triplet counts per anchor due to skipped comparisons, we report anchor-balanced agreement, by weighting the triplets $t \in \mathcal{T}_u$ of each anchor $u \in \mathcal{A}$ equally:
\begin{equation}
\text{Bal. Agr.} = \frac{1}{|\mathcal{A}|} \sum_{u \in \mathcal{A}} \frac{1}{|\mathcal{T}_u|} \sum_{t \in \mathcal{T}_u} \mathbf{1}[\Delta_t > 0].
\end{equation}
We also report micro agreement $\frac{1}{|\mathcal{T}|}\sum_{t \in \mathcal{T}} \mathbf{1}[\Delta_t > 0]$, which weights all triplets equally. Treating the prediction as a binary classification task, we compute macro-F1 as the average of the per-class F1 scores, where each F1 is the harmonic mean of precision and recall. Finally, Cohen's $\kappa = (p_o - p_e)/(1 - p_e)$ measures chance-corrected agreement, with $p_o$ the observed micro agreement and $p_e$ the expected agreement under independence of predictions and labels.

\subsection{Experimental Settings}
For the text modality, we query \texttt{gpt-oss-120b} with 68{,}913 triplets constructed from expert descriptions. We fit SOE embeddings to the resulting constraints using Adam with AMSGrad, learning rate $0.05$, batch size $2048$, and hinge margin $\delta = 0$. We train for $50$ epochs with anchor-balanced sampling, which resamples triplets each epoch to give equal weight to all anchors regardless of their triplet count. The embedding dimension is set to $4$ and we ablate key settings in \secref{sec:ab}.

The vision model predictor head is trained for $50$ epochs with a batch size of $32$, a learning rate of $10^{-3}$, a weight decay of $10^{-5}$, and a cosine learning rate schedule. The VICReg hyperparameters $\lambda$, $\mu$, and $\nu$ are set to their default values of $25$, $25$, and $1$, respectively. The embedding dimension is set to $512$. Input images are resized and non-empty randomly cropped to a resolution of $224 \times 224$. Data augmentations comprise random horizontal and vertical flips, shifts, scaling, rotations, color jitter, brightness and contrast adjustments, Gaussian noise, and Gaussian blur. At inference time, images are resized to $224 \times 224$.

\begin{table}[t]
\centering
\caption{Agreement metrics for text, vision, and fused embeddings. Balanced anchor agreement, micro agreement, and macro-F1 are reported in \%; Cohen's $\kappa$ is unitless.}
\setlength{\tabcolsep}{4pt}
\footnotesize
\begin{tabular}{@{}llcccc@{}}
\toprule
\textbf{Modality} & \textbf{Method} & \textbf{Bal. Agr.}$\uparrow$ & \textbf{Micro}$\uparrow$ & \textbf{Macro-F1}$\uparrow$ & \textbf{$\kappa$}$\uparrow$ \\
\midrule
\multirow{6}{*}{\textit{TEXT}}
 & MiniLM~\cite{wang2020minilm} & 56.9 & 58.0 & 57.7 & 0.155 \\
 & gpt-oss-120b mean~\cite{openai2025gptoss120b} & 58.9 & 58.2 & 58.0 & 0.160 \\
 & MedGemma-27B mean~\cite{medgemma2025} & 61.3 & 61.8 & 61.5 & 0.231 \\
 & PubMedBERT~\cite{gu2021pubmedbert} & 61.7 & 61.4 & 61.4 & 0.230 \\
 & RoBERTa~\cite{liu2019roberta} & 63.7 & 63.3 & 63.1 & 0.262 \\
 & \net{} (Text only) & \textbf{67.4} & \textbf{66.1} & \textbf{66.1} & \textbf{0.321} \\
\midrule
\multirow{7}{*}{\textit{VISION}}
 & MedGemma-27B~\cite{medgemma2025} & 50.3 & 52.4 & 52.2 & 0.044\\
 & SigLIP~\cite{zhai2023sigmoid} & 56.3 & 58.0 & 57.9 & 0.165 \\
 & SAM~\cite{kirillov2023segment} & 57.4 & 56.7 & 56.6 & 0.138 \\
 & PanDerm~\cite{panderm2024} & 58.4 & 58.0 & 58.0 & 0.163 \\
 & DINOv3~\cite{dinov3} & 66.9 & 67.2 & 67.1 & 0.350 \\
 & DermLiP~\cite{dermlip2024} & 67.9 & 63.1 & 62.9 & 0.258 \\
 & \net{} (Vision only) & \textbf{71.6} & \textbf{70.2} & \textbf{70.1} & \textbf{0.403} \\
\midrule
\multirow{2}{*}{\textit{FUSED}}
 & \net{} (similarity) & 72.5 & 72.1 & 72.1 & 0.442 \\
 & \net{} (uncertainty) & \textbf{73.5} & \textbf{72.3} & \textbf{72.3} & \textbf{0.446} \\
\bottomrule
\end{tabular}
\label{tab:comparison}
\end{table}

\subsection{Results}
We compare against relevant baselines for each modality. For text, we consider ROBERTa~\cite{liu2019roberta}, PubMedBERT~\cite{gu2021pubmedbert}, MiniLM~\cite{wang2020minilm}, and MedGemma-27B~\cite{medgemma2025}. For gpt-oss-120b~\cite{openai2025gptoss120b} and MedGemma-27B, we create representations using mean pooling over token embeddings.
For vision, we benchmark against SigLIP~\cite{zhai2023sigmoid}, PanDerm~\cite{panderm2024}, SAM~\cite{kirillov2023segment}, MedGemma~\cite{medgemma2025}, DINOv3~\cite{dinov3}, and DermLIP~\cite{dermlip2024}. For all baselines, we use the respective training image size and perform mean pooling of marked wound features.

\tabref{tab:comparison} summarizes the performances. For text, SOE yields a gain of 3.7 percentage points (pp) over the strongest baseline (RoBERTa), demonstrating the benefit of encoding medical knowledge into an embedding space via triplet structure. For vision, our proposed wound attention pooling with non-contrastive learning surpasses the best baseline (DermLIP) by 3.7~pp, indicating it captures clinically relevant wound features more effectively. Notably, MedGemma achieves 61.3\% on text (above baseline) but only 50.3\% on vision (random chance), indicating its visual representations lack discriminative information for this task despite reasonable text performance.
Vision and text capture complementary information: uncertainty-aware fusion reaches 73.5\%, a gain of 1.9~pp over vision alone (71.6\%) and 6.1~pp over text alone (67.4\%), demonstrating that the modalities provide independent signal. Similarity-based fusion achieves 72.5\%, revealing that cross-modal agreement filtering can improve upon unimodal.
We report balanced accuracy, micro accuracy, macro-F1, and Cohen's $\kappa$, which handle class imbalance and chance agreement differently. All metrics yield comparable rankings, supporting the absence of data imbalance artefacts. On the Landis--Koch scale, fusion achieves $\kappa = 0.45$ (moderate agreement), compared to $\kappa \approx 0.32$--$0.40$ for unimodal methods.


\begin{figure}[t]
\centering
\setlength{\tabcolsep}{2pt}
\newcolumntype{M}[1]{>{\centering\arraybackslash}m{#1}}
\small
\begin{tabular}{M{0.3cm}M{0.18\textwidth}M{0.18\textwidth}M{0.18\textwidth}M{0.18\textwidth}M{0.18\textwidth}}
& Ref. & \multicolumn{2}{c}{Most Similar} & \multicolumn{2}{c}{Most Dissimilar} \\
(i) & \includegraphics[width=\linewidth]{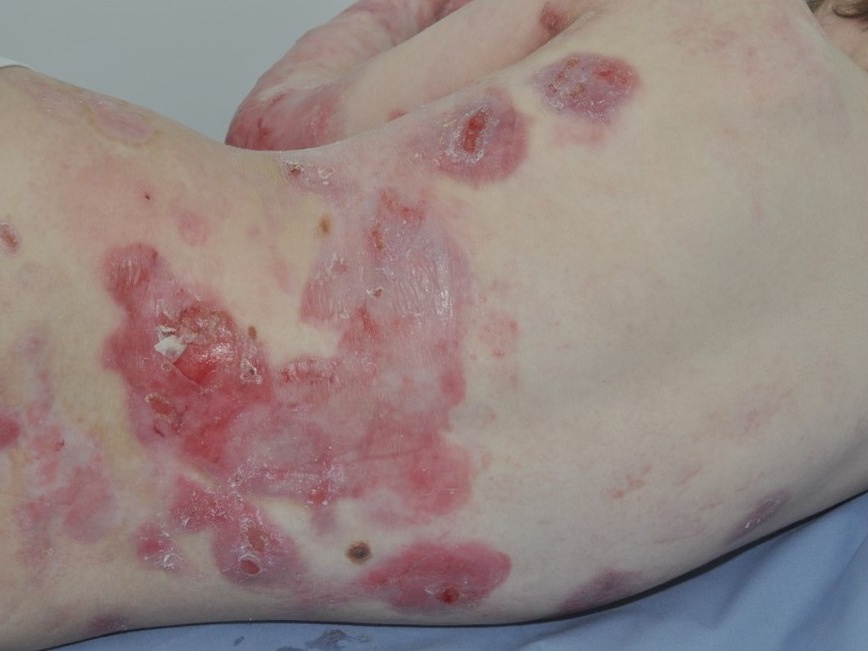} & \includegraphics[width=\linewidth]{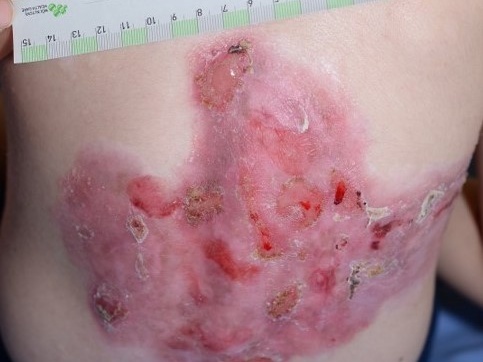} & \includegraphics[width=\linewidth]{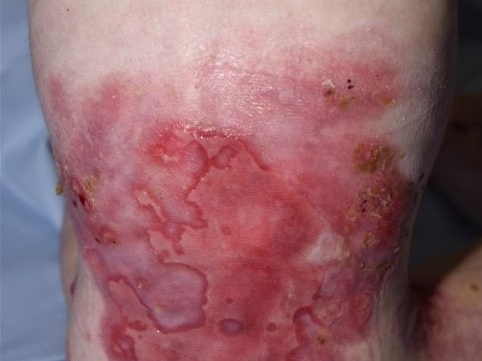} & \includegraphics[width=\linewidth]{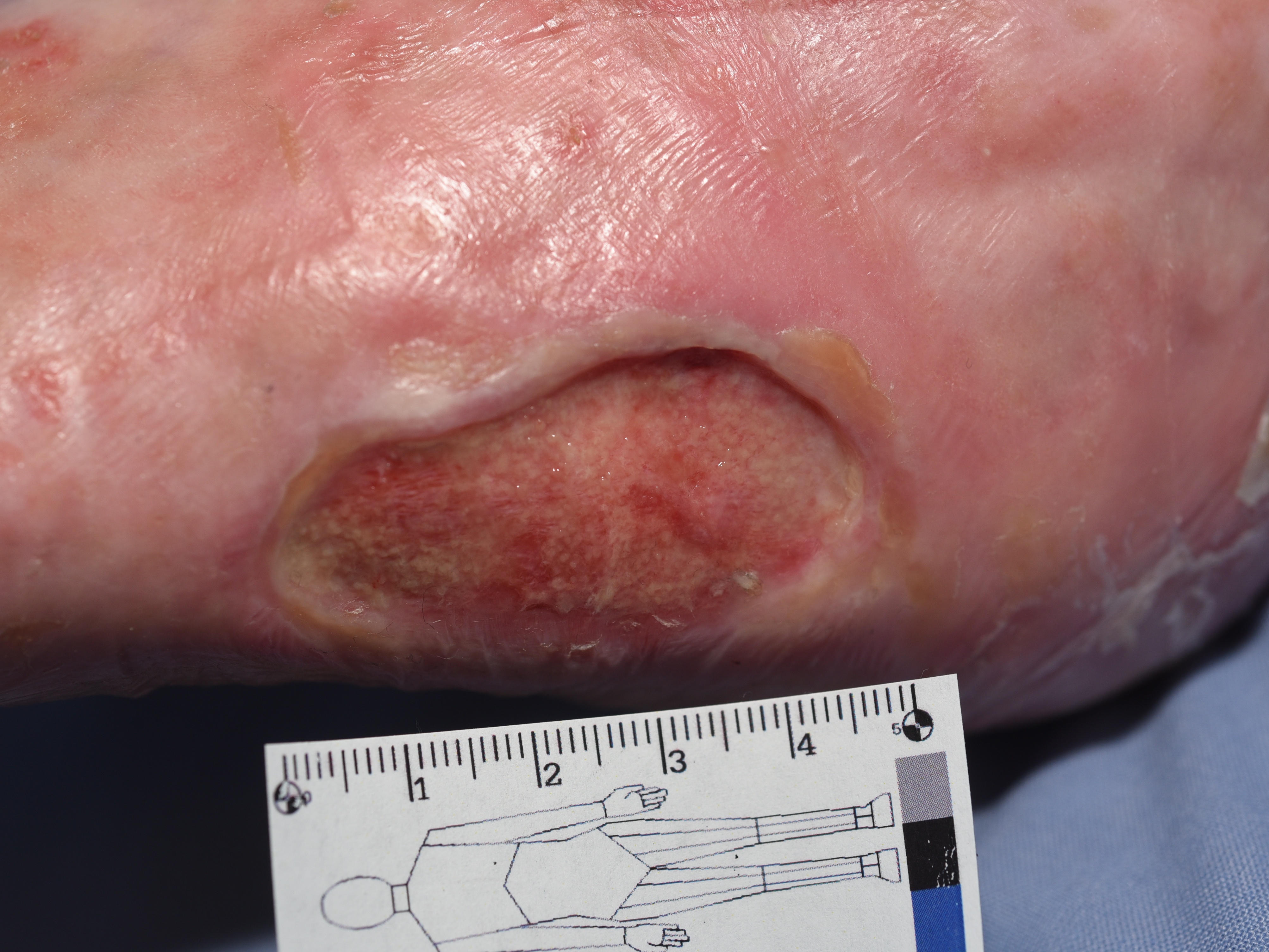} & \includegraphics[width=\linewidth]{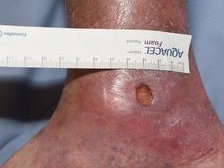}\\[2pt]
(ii) & \includegraphics[width=\linewidth]{figures/qual/60.jpg} & \includegraphics[width=\linewidth]{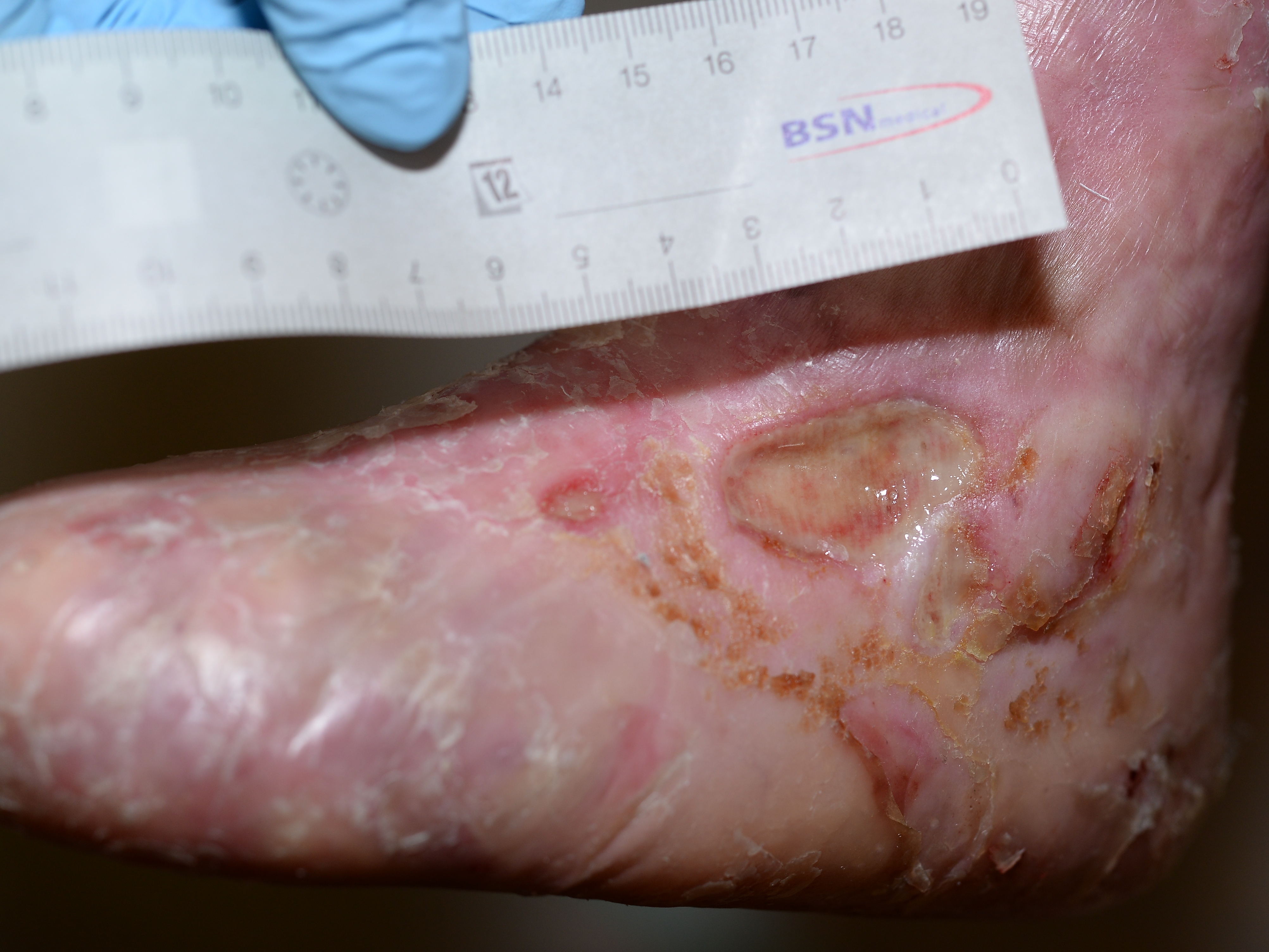} & \includegraphics[width=\linewidth]{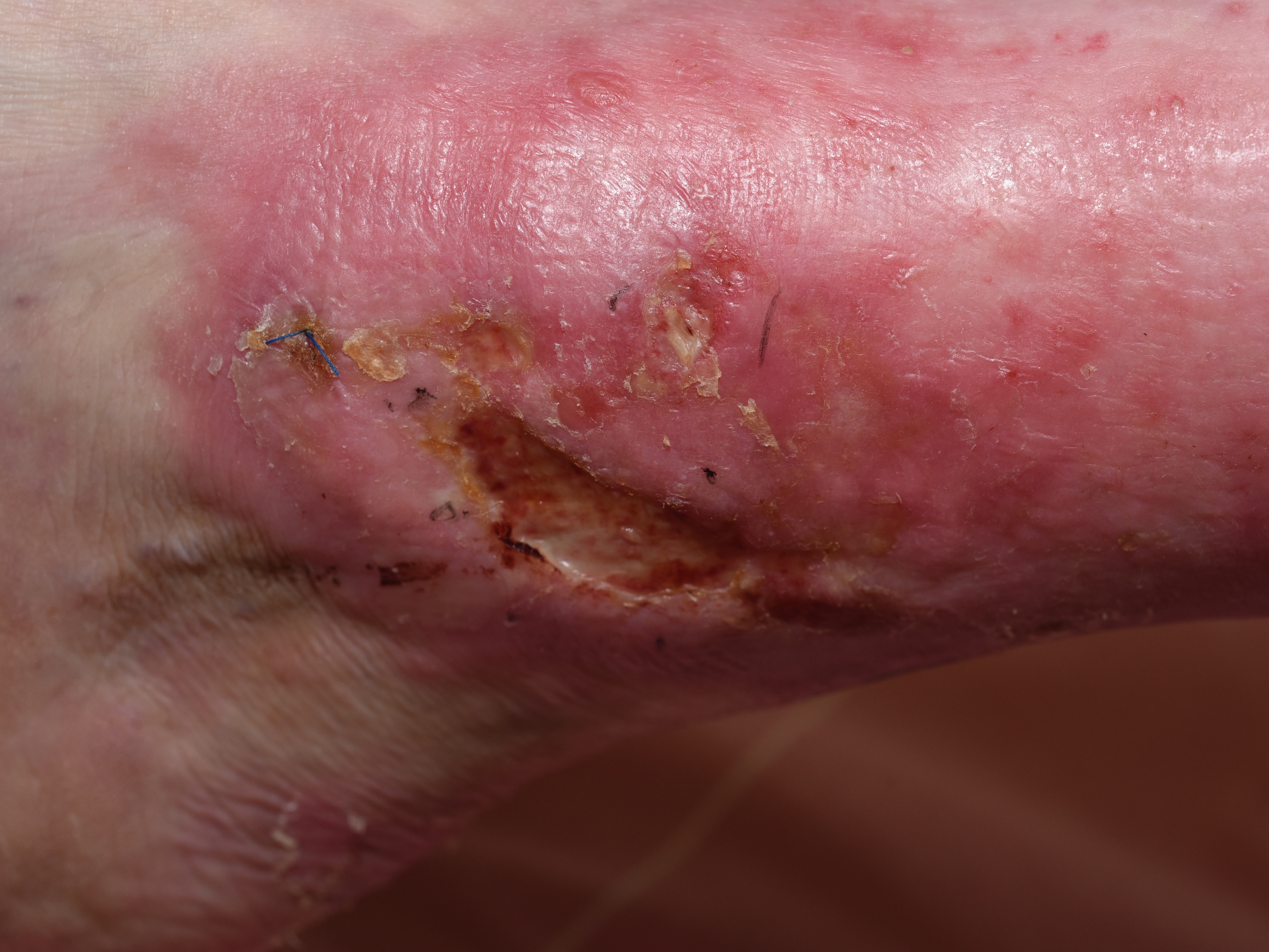} & \includegraphics[width=\linewidth]{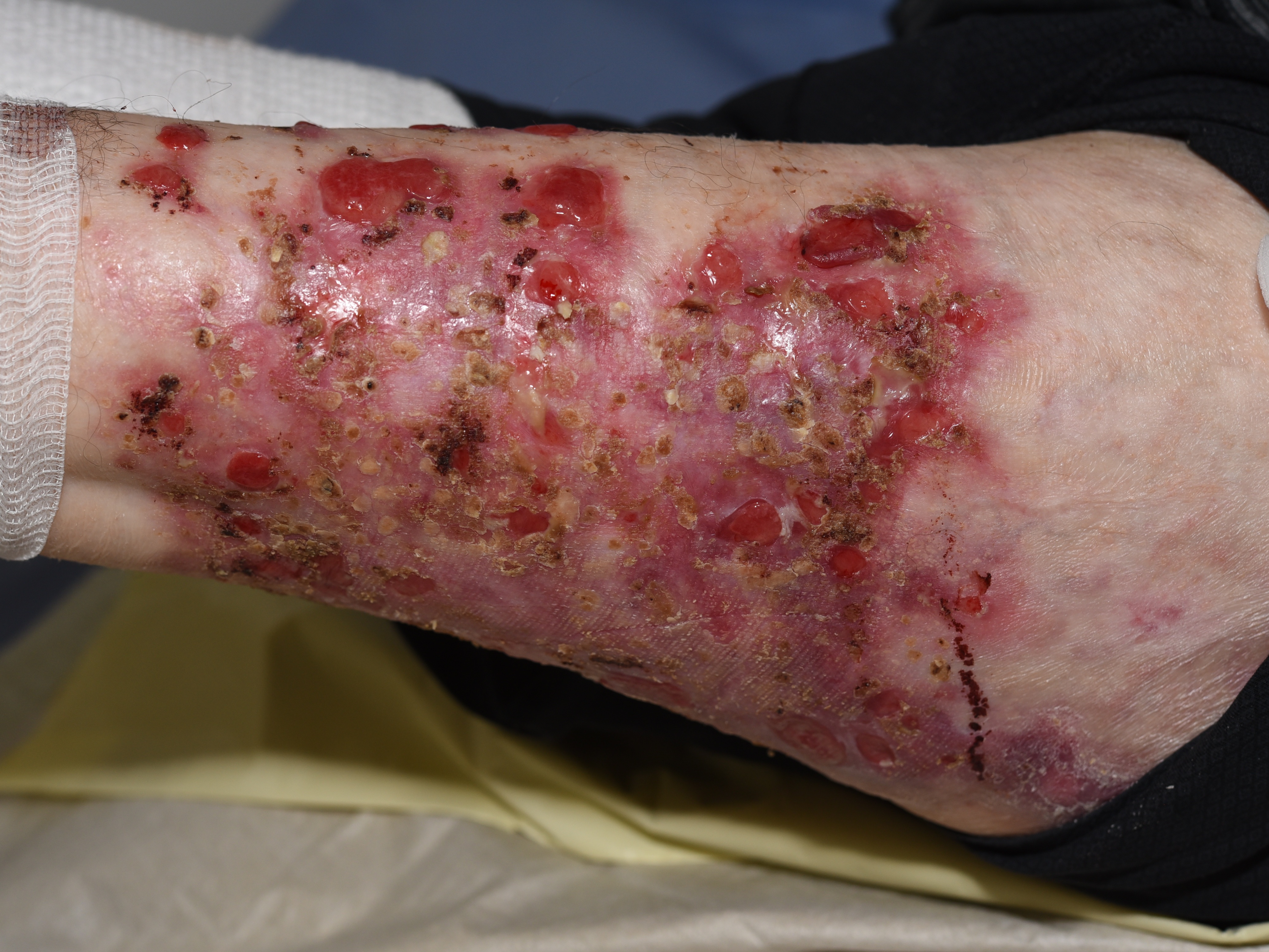} & \includegraphics[width=\linewidth]{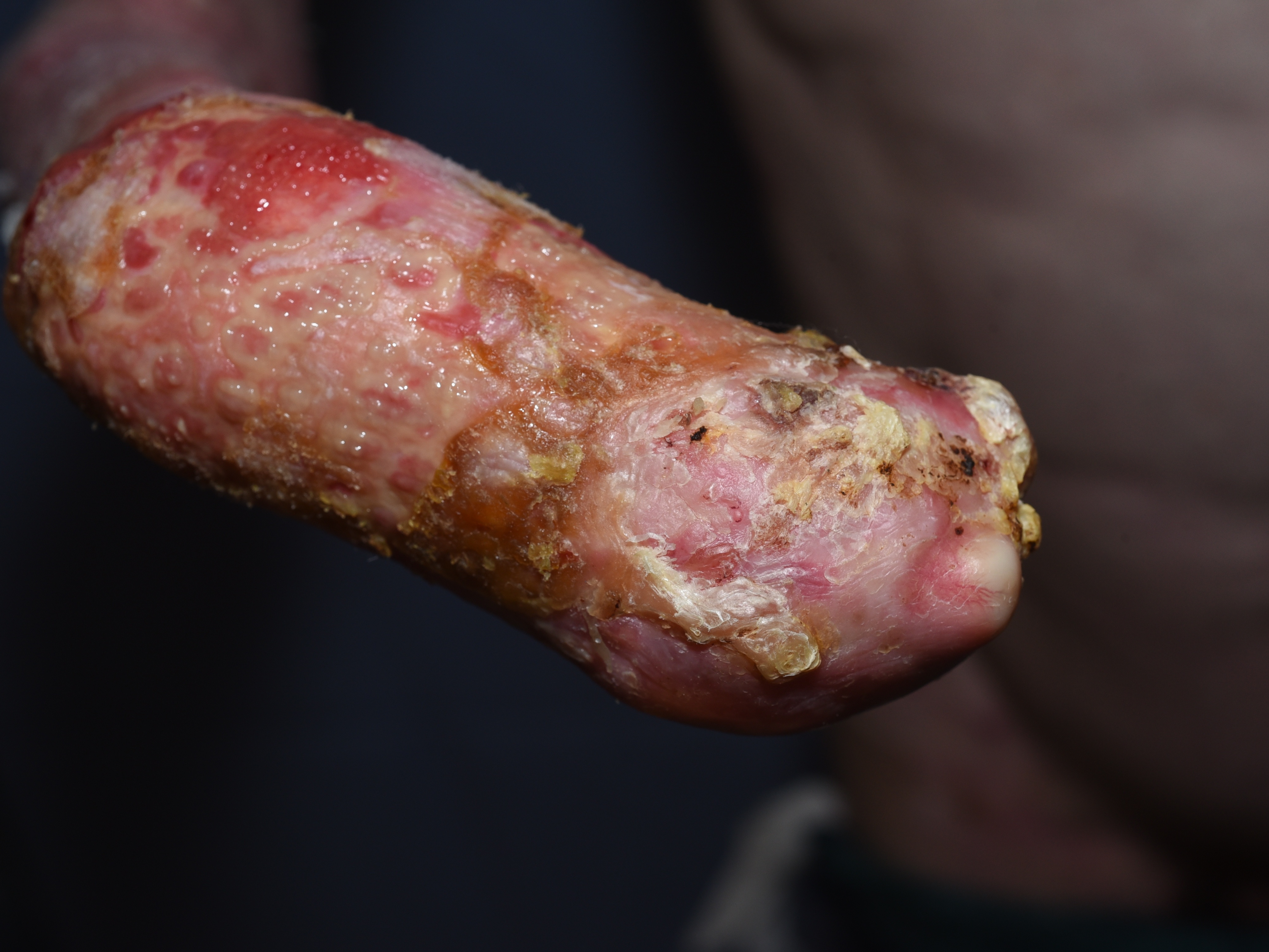}\\
\end{tabular}
\caption{Nearest neighbor retrieval. In (i), similar wounds show superficial granulating lesions; dissimilar are deep ulcers. In (ii), similar wounds are deep fibrotic ulcers; dissimilar are superficial and inflamed.}
\label{fig:qual}
\end{figure}

We illustrate in \figref{fig:qual} that the exhibited embedding similarities of \net{} (variance) are medically meaningful. Similar images have $d < 0.35$, while dissimilar images have $d > 0.8$. Row (i) groups superficial lesions with granulation tissue, contrasting with deep ulcers. Row (ii) groups deep fibrotic ulcerations, contrasting with superficial inflamed wounds—demonstrating that the embedding captures wound depth, granulation, and inflammation.

\subsection{Ablation}\label{sec:ab}

\tabref{tab:text} reports SOE ablations over embedding dimensionality, triplet budget and oracle models.
Embedding dimensionality exhibits an optimum at $D=4$ (67.4\%), with performance degrading for both lower and higher dimensions.
Very low-dimensional embeddings ($D=2$) lack the capacity to capture the underlying similarity structure, while higher-dimensional spaces overfit to noise in the LLM's triplet judgments.
Regarding the triplet budget, accuracy improves monotonically from 55.2\% at 0.1\% of triplets to 67.4\% with the full set, indicating that using all triplet constraints helps uncover the LLM's latent similarity structure.
At very small budgets (0.1\%), accuracy degrades to slightly above chance (50\%), confirming the need for sampling a larger subspace of triplet questions.
We also compare different LLMs as triplet oracles: gpt-oss-120b (67.4\%) outperforms both gpt-oss-20b (61.7\%) and the medical-domain MedGemma-27B (61.1\%), which even underperforms compared to mean token embeddings (61.3\%). This suggests that general reasoning capability matters more than domain-specific fine-tuning for pairwise similarity judgments.


\tabref{tab:vision} reports the effect of different self-supervised adaption strategies. Attention pooling over wound regions outperforms mean pooling with the same predictor head by $2.3$ pp, because it more effectively weights characteristic wound features. Regarding different self-supervised loss functions, the triplet loss achieves its best performance with a batch size of 8, while contrastive learning benefits from a larger batch size of 128 and a higher learning rate of 0.001, consistent with the known sensitivity of contrastive objectives to the number of in-batch negatives~\cite{chen2020simple}. Finally, the non-contrastive method VICReg~\cite{bardes2021vicreg} outperforms both alternatives under this low-data regime, which we attribute to its independence from negative pairs — a property that confers a significant advantage when the training set is small.

\IfFileExists{generated/tab_soe_ablation.tex}{

\begin{table}[t]
\centering
\begin{minipage}[t]{0.48\textwidth}
\centering
\vspace{0pt}
\setlength{\tabcolsep}{3pt}
\footnotesize
\begin{tabular}{@{}ccccc@{}}
\toprule
\multicolumn{5}{c}{Embedding Dim.} \\
\midrule
$2$ & $3$ & $4$ & $5$ & $6$ \\
62.6 & 65.0 & \textbf{67.4} & 64.2 & 64.9\\
\midrule
\multicolumn{5}{c}{No.\ Triplets} \\
\midrule
0.1\% & 1\% & 10\% & 100\% & \\
55.2 & 62.6 & 63.6 & \textbf{67.4} & \\
\midrule
\multicolumn{5}{c}{LLM Oracle} \\
\midrule
MedG & GPT-20 & GPT-120 & & \\
61.1 & 61.7 & \textbf{67.4} & & \\
\bottomrule
\end{tabular}
\vspace{0.3em}
\subcaption{Text adaption}
\label{tab:text}
\end{minipage}
\hfill
\begin{minipage}[t]{0.45\textwidth}
\centering
\vspace{0pt}
\setlength{\tabcolsep}{3pt}
\footnotesize
\begin{tabular}{l c}
\toprule
\multicolumn{2}{c}{Feature Pooling}\\
\midrule
Mean & 69.3 \\
Att. (ours) & \textbf{71.6} \\
\midrule
\multicolumn{2}{c}{Loss Function} \\
\midrule
Contrastive~\cite{khosla2020supervised} & 68.1 \\
Triplet~\cite{schroff2015facenet} & 70.1 \\
VICReg (ours) & \textbf{71.6} \\
\bottomrule
\end{tabular}
\vspace{0.3em}
\subcaption{Vision Adaption}
\label{tab:vision}
\end{minipage}
\caption{Ablation studies. (a) \net{} SOE ablations using GPT-120B as oracle. (b)~Vision model pooling and loss variations. Bal.\ Agr. recorded in [\%].}
\label{tab:ablations}
\vspace{-2em}
\end{table}

}{
\noindent\textit{SOE ablation table pending artifact generation (\texttt{generated/tab\_soe\_ablation.tex}).}
}


\section{Conclusion}




Our findings suggest that visual and text foundation models encode some clinically relevant structure, but miss fine-grained distinctions for RDEB wounds which experts consider important. Targeted adaptation, i.e. VICReg fine-tuning for vision and soft ordinal embedding for LLM triplet judgments, improves this alignment using only 53 unlabeled images. A gap with expert judgment remains, yet multimodal fusion narrows this discrepancy by combining complementary information from vision and text. Our proposed triplet-based evaluation is effective when labeled data is scarce but expert similarity judgments are obtainable.



\subsubsection*{Funding}
Funded by the Deutsche Forschungsgemeinschaft (DFG, German Research Foundation) -- Project-ID 499552394 -- SFB 1597.

\subsubsection*{Conflict of Interest}
The authors have no competing interests to declare that are relevant to the content of this article.

\newpage
\bibliographystyle{splncs04}
\bibliography{bibliography,rdeb_embedding}

\end{document}